\documentclass[10pt,twocolumn,letterpaper]{article}

\usepackage{cvpr}
\usepackage{times}
\usepackage{epsfig}
\usepackage{graphicx}
\usepackage{amsmath}
\usepackage{amssymb}
\usepackage{stfloats}
\usepackage{array}
\usepackage{booktabs}
\usepackage{cite}

\usepackage{color}
\usepackage{subcaption}
\usepackage{algorithm}
\usepackage[noend]{algpseudocode}
\usepackage{color, colortbl}
\usepackage{xspace, xcolor}

\definecolor{Gray}{gray}{0.9}

\usepackage{multirow}
\usepackage{multicol}
\usepackage{tabularx}
\newcolumntype{Y}{>{\centering\arraybackslash}X}

\newcommand{\comment}[1]{}

\usepackage[pagebackref=true,breaklinks=true,letterpaper=true,colorlinks,bookmarks=false]{hyperref}

\cvprfinalcopy 

\def\cvprPaperID{4882} 

\ifcvprfinal
\fi
\begin{document}

\title{Dance with Flow: Two-in-One Stream Action Detection}
\author{Jiaojiao Zhao and
Cees G. M. Snoek\\
University of Amsterdam\\
}
\maketitle
\thispagestyle{empty}

\begin{abstract}
The goal of this paper is to detect the spatio-temporal extent of an action. The two-stream detection network based on RGB and flow provides state-of-the-art accuracy at the expense of a large model-size and heavy computation. We propose to embed RGB and optical-flow into a single two-in-one stream network with new layers. A motion condition layer extracts motion information from flow images, which is leveraged by the motion modulation layer to generate transformation parameters for modulating the low-level RGB features. The method is easily embedded in existing appearance- or two-stream action detection networks, and trained end-to-end. Experiments demonstrate that leveraging the motion condition to modulate RGB features improves detection accuracy. With only half the computation and parameters of the state-of-the-art two-stream methods, our two-in-one stream still achieves impressive results on {\it{UCF101-24}}, {\it{UCFSports}} and {\it{J-HMDB}}.
\end{abstract}

\section{Introduction}
This paper strives for the spatio-temporal detection of human actions in video, which is a crucial ability for self-driving cars, autonomous care robots, and advanced video search engines. The leading approach for this challenging problem relies on fast detectors at the frame level \cite{peng2016multi,singh2017online}, which are then linked \cite{gkioxari2015finding,singh2017online,behl2017incremental} or tracked \cite{weinzaepfel2015learning} over time. Kalogeiton \etal~\cite{kalogeiton2017action} and Singh \etal~\cite{singh2018tramnet} further showed it is advantageous to stack the features from subsequent frames before predicting action class scores and determining the enclosing tube. Most of the state-of-the-art action detectors exploit a two-stream architecture~\cite{simonyan2014two}, one for RGB and one for optical-flow, which are individually trained before fusion. However, the double computation and parameter demand of two-stream methods does not lead to double accuracy compared to a single stream. We propose to embed RGB and optical-flow into a single stream for action detection.

\begin{figure}
\centering
\includegraphics[scale=0.55]{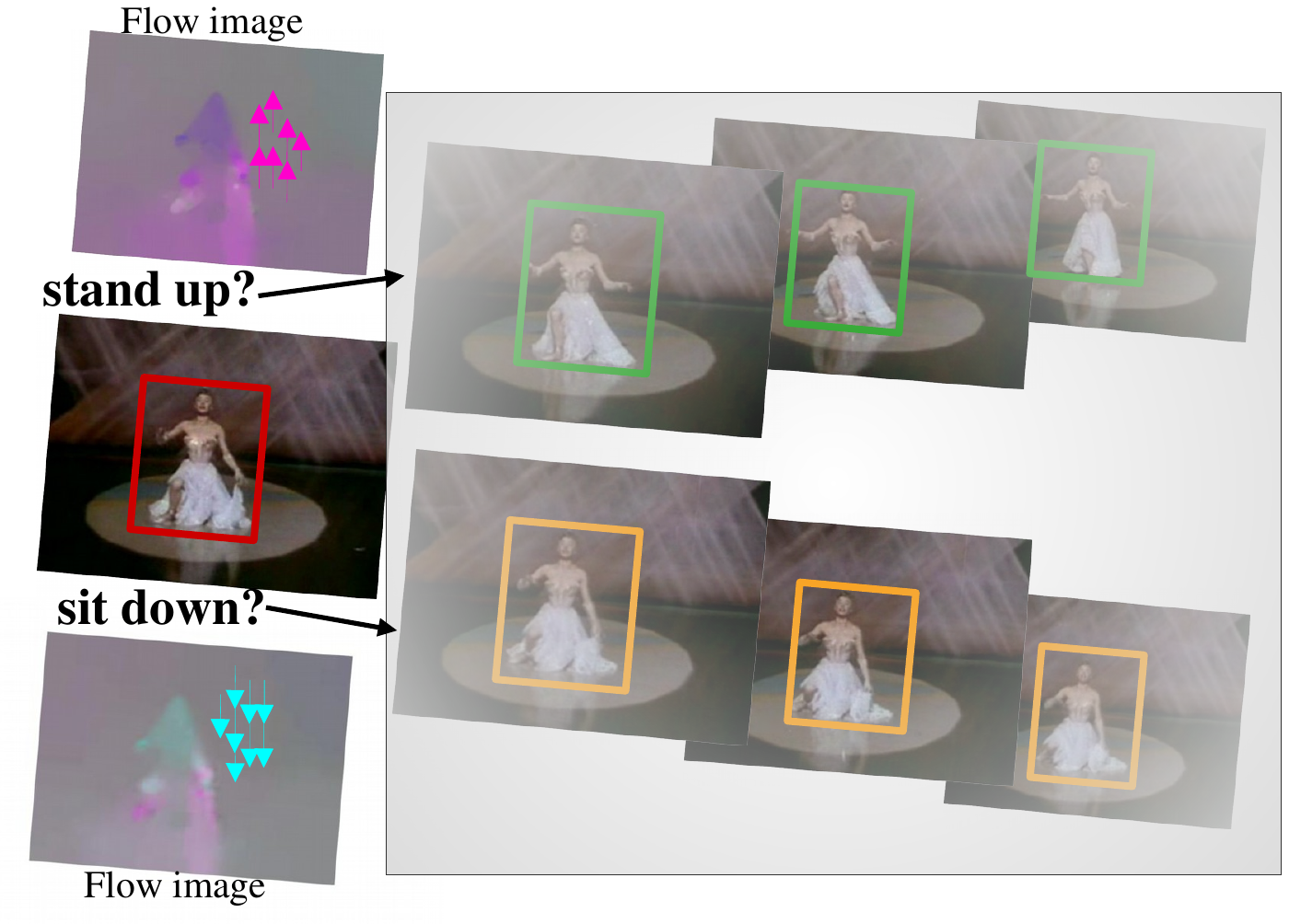}
\caption{\textbf{Two-in-one stream.} We propose to embed RGB and optical-flow into a single stream for spatio-temporal action detection. Besides efficiency gains, it helps recognizing whether the dancer in the current frame is standing up or sitting down without considering the future. By utilizing information from flow images, the dancer is given a moving direction, up or down, better indicating the action.}
\label{Fig:motivation1}
\end{figure}

We are inspired by progress on feature normalization, especially conditional normalization~\cite{ioffe2015batch,dumoulin2017learned,ghiasi2017exploring}, which has been successfully employed to visual question answering~\cite{de2017modulating}, visual reasoning~\cite{perez2017film}, image style transfer~\cite{huang2017arbitrary} and super-resolution~\cite{wang2018recovering}. Peretz \etal~\cite{perez2017film} propose a feature-wise linear modulation layer which enables a recurrent neural network over an input question to influence convolutional neural network computation over an image. It demonstrates that features are capable to be modulated via a simple feature-wise affine transformation based on conditioning information. However, as their modulation layer is agnostic to spatial location, it is unsuited for action detection. In~\cite {wang2018recovering}, Wang \etal developed a  spatial feature transform layer, which is conditioned on categorical semantic probability maps, to modulate a super-resolution network.  Encouraged by these works, we propose a motion condition layer and a motion modulation layer to adjust an RGB-stream for spatio-temporal action detection.

We make the following contributions in this paper. We propose to embed RGB and optical-flow into a single stream for spatio-temporal action detection. It reduces the computational costs of conventional two-stream detection networks by half while maintaining its high accuracy. We introduce the two-in-one stream with motion condition layer and motion modulation layer, which learns video representations of appearance-stream features conditioned on optical-flow. As shown in Figure~\ref{Fig:motivation1}, the motion condition will guide the model to pay more attention on what moves, rather than the static background. The method is easily embedded in existing appearance- or two-stream action detection networks, and trained end-to-end, leading to new state-of-the-art on {\it{UCF101-24}}, {\it{UCFSports}} and {\it{J-HMDB}}.

\section{Related Work}
The spatio-temporal detection of human actions in video has a long tradition in computer vision, \eg~\cite{dalal2006human,klaser2008spatio,cao2010cross}. Early success came from detection based on exhaustive cuboid search, efficient feature representations, and SVM-based learning, \cite{yuan2011discriminative,tran2012max,tian2013spatiotemporal}. This was later extended with more flexible sequences of bounding boxes \cite{lan2011discriminative, tran2014video,yu2015fast}, or spatio-temporal proposals~\cite{van2015apt,JainIJCV17}, together with engineered appearance and motion features, most notably the dense trajectories~\cite{oneata2014efficient}. The past few years, architectures integrating detection and deep representation learning have been leading~\cite{sun2018actor,girdhar2018better,yang2017spatio,hou2017tube,li2018recurrent,xie2018rethinking,duarte2018videocapsulenet}, typically combining appearance and flow streams~\cite{singh2017online,kalogeiton2017action,gu2017ava,he2018generic}. We follow this tradition.

The two-stream network was first introduced by Simonyan and Zisserman in~\cite{simonyan2014two}. Their convolutional architecture included a separate RGB-stream and a flow-stream, which were combined by late fusion, for SVM-based action classification. In ~\cite{feichtenhofer2016convolutional}, Feichtenhofer \etal investigated a number of ways to fuse the RGB and flow streams in order to best take advantage of their fused representation for action classification. While we concentrate on action detection in the paper, we are interested in RGB and flow as well, but rather than combining the two streams in a late fusion, we prefer a single stream.


Gkioxari and Malik~\cite{gkioxari2015finding} introduced a two-stream architecture with R-CNN detectors in action detection. They fused features from the last layer of an RGB- and a flow-stream, and then trained action specific SVM classifiers. A Viterbi algorithm~\cite{vsramek2007line} was adopted to link the detection boxes per frame into a tube. Weinzaepfel \etal~\cite{weinzaepfel2015learning} also used a two-stream R-CNN detector but replaced the linking by a tracking-by-detection method. Both methods are not end-to-end trainable and restricted to trimmed videos.


End-to-end two-stream detectors based on faster-RCNN were proposed in~\cite{peng2016multi,saha2016deep}. In~\cite{peng2016multi}, Peng and Schmid performed region of interest pooling and score fusion to incorporate an RGB-stream and a flow-stream. In ~\cite{hou2017tube}, Hou \etal extended 2D region of interest pooling to 3D tube-of-interest pooling with 3D convolutions, which directly generate tubelets for action detection. Singh \etal adopted a two-stream single-shot-multibox detector (SSD)\cite{liu2016ssd} for realizing real time detection in~\cite{singh2017online}. Singh \etal~\cite{singh2018tramnet} also introduced a transition matrix to generate a set of action proposals on pairs of frames. Kalogeiton \etal~\cite{kalogeiton2017action} proposed to exploit temporal continuity by taking as much as six frames as input for their single-shot multibox detector, leading to state-of-the-art results. In this paper, we take the single-shot multibox detector network as our backbone, using single~\cite{singh2017online} or multiple~\cite{kalogeiton2017action} frames as input, but rather than separating the streams for RGB and flow we introduce a single two-in-one stream.

Li \etal~\cite{li2018videolstm} proposed an action detector using an LSTM architecture with motion-based attention. Our two-in-one stream not only takes motion as attention, which helps to locate actions, but also uses motion to modulate RGB features which helps to better classify actions. Moreover, our method is easily embedded in existing appearance- or two-stream action detection and classification networks.

\begin{figure*}[t!]
\centering
\includegraphics[scale=0.60]{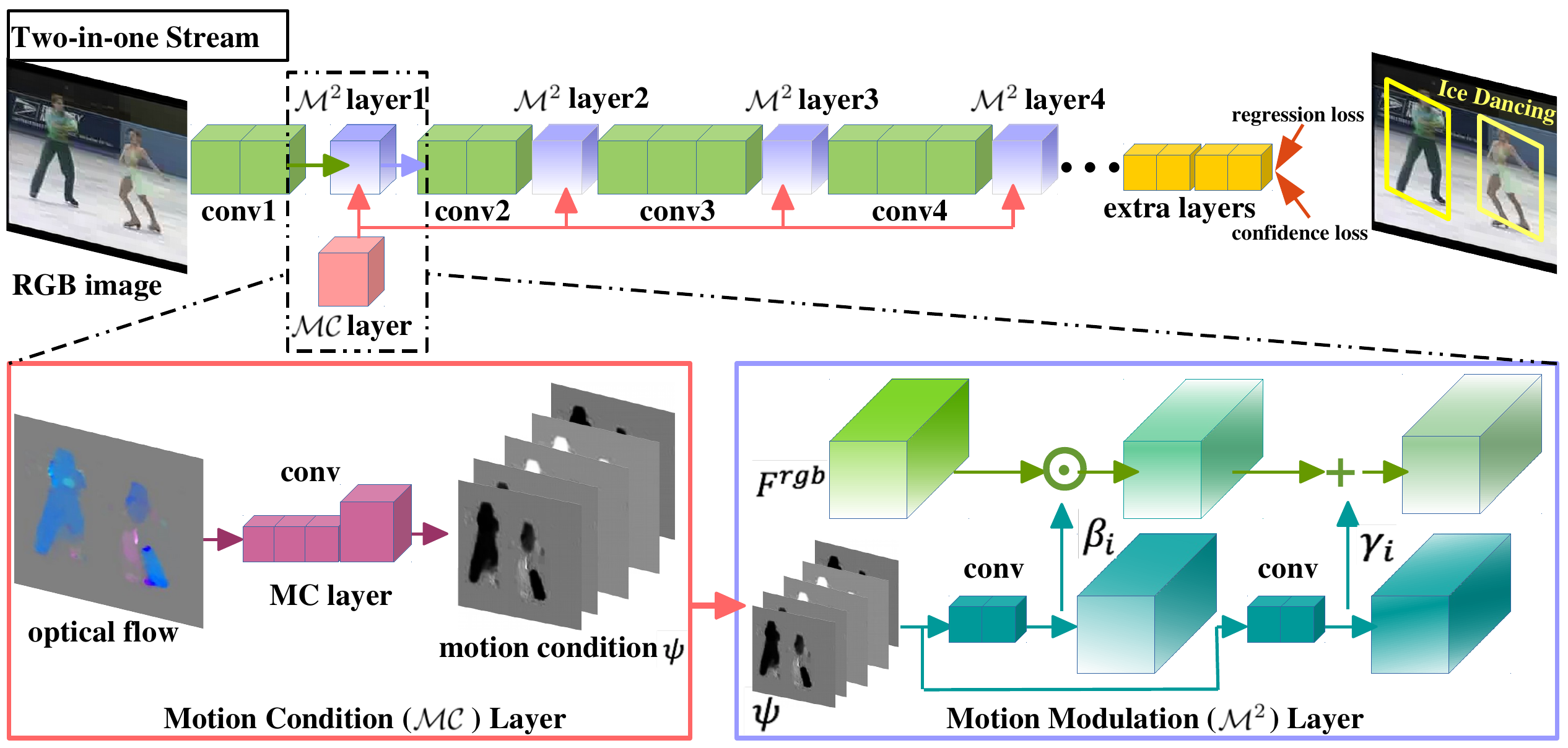}
\caption{\textbf{Two-in-one network architecture}. The motion condition layer (pink cube) maps flow images to prior condition information. The condition inputs to the motion modulation layer (purple cube) to generate transformation parameters which are used to modulate RGB features ($F^{rgb}$). The network has half the computation and parameters of a two-stream equivalent, while obtaining better action detection accuracy.
}
\label{Fig:network}
	\vspace{-5pt}
\end{figure*}
\section{Two-in-One Network}
We define the RGB-stream network $\mathcal{D}^{rgb}_{\theta}$ trained on single frame for spatio-temporal action detection as:
\begin{equation}
(L^{rgb}, S^{rgb}) = \mathcal{D}^{rgb}_{\theta}(I^{rgb})
\end{equation}
where $I^{rgb}\in\mathbb{R}^{H \times W \times 3}$ is a single RGB frame of height $H$ and width $W$ which is the input for the network $\mathcal{D}^{rgb}_{\theta}$. $L^{rgb}\in\mathbb{R}^{Q \times 4}$ and $S^{rgb}\in\mathbb{R}^{Q \times (P+1)}$ are $Q$ box locations and corresponding box classification scores for $P$ action classes and a background class. $\theta$ represents the parameters of the learned network. Similarly, we define a flow-stream network on single frame for spatio-temporal action detection as:
\begin{equation}
(L^{of}, S^{of}) = \mathcal{D}^{of}_{\theta}(I^{of})
\end{equation}
$I^{of}\in\mathbb{R}^{H \times W \times 2}$ is a single optical flow image with $x$ and $y$ components of the velocity respectively in two channels. The two-stream method includes training the two networks $\mathcal{D}^{rgb}_{\theta}$ and $\mathcal{D}^{of}_{\theta}$ independently, and fuses the results $(L^{rgb}, S^{rgb})$ and  $(L^{of}, S^{of})$.

{\bf{Motion condition layer.}} In our method, $I^{of}$ is regarded as a motion map with the same resolution as the corresponding RGB image $I^{rgb}$. We take $I^{of}$ as prior information $\Psi$ when applying an RGB-stream network $\mathcal{D}^{rgb}_{\theta}$ to estimate where and what actions may occur. Then we formulate our two-in-one network as a condition network:

\begin{equation}
\begin{aligned}
(L^{{\twoheadrightarrow}}, S^{{\twoheadrightarrow}}) &= \mathcal{D}^{{\twoheadrightarrow}}_{\theta}(I^{rgb}|\Psi)\\
&=\mathcal{D}^{{\twoheadrightarrow}}_{\theta}(I^{rgb}|\mathcal{MC}(I^{of}))
\end{aligned}
\end{equation}
\begin{equation}
\Psi = \mathcal{MC}(I^{of}) = \mathcal{MC}((I^{of_x},I^{of_y}))
\end{equation}

${\twoheadrightarrow}$ means two-in-one stream, ${\mathcal{MC}(\cdot)}$ is a mapping function to generate simple features from the flow images. So the two-in-one stream $\mathcal{D}^{{\twoheadrightarrow}}_{\theta}$ learns a model conditioned on motion information by a motion condition layer.

{\bf{Motion modulation layer.}}
We introduce a motion modulation ($\mathcal{M}^2$) layer to modify the features learned from RGB images. An $\mathcal{M}^2$ layer is able to influence the appearance network by incorporating motion and weighting the action area. We first learn a pair of affine transformation parameters ($\beta, \gamma$) from the prior flow condition $\Psi$ by a function $\mathcal{F}: \Psi\longmapsto(\beta, \gamma)$. Concretely, the two-in-one network is further expressed as:
\begin{equation}
\begin{aligned}
(\beta, \gamma) &= \mathcal{F}(\Psi),\\
(L^{{\twoheadrightarrow}}, S^{{\twoheadrightarrow}}) &= \mathcal{D}^{{\twoheadrightarrow}}_{\theta}(I^{rgb}|\beta, \gamma)
\end{aligned}
\end{equation}
In order to modulate the appearance network, we apply a transform function $\mathcal{M}^2(\cdot)$  with the learned transformation parameters ($\beta, \gamma$) to the RGB features $F^{rgb}$.
\begin{equation}
\mathcal{M}^2(F^{rgb}) = \beta\odot{F^{rgb}}+\gamma
\end{equation}
$\odot$ is an element-wise multiplication operation. The RGB feature maps $F^{rgb}$ has the same dimensions with parameters $\beta$ and $\gamma$. The flow information represented by ($\beta, \gamma$) influences the appearance network by both feature-wise and spatial-wise manipulations. The complete network with the motion condition layer and the motion modulation layer is shown in Figure~\ref{Fig:network}.


{\bf{Network architecture.}}  Due to sparsity of flow images, we adopt simple convolutional layers to extract low-level motion condition information. $1\times1$ convolutional layers attempt to keep the spatial pixel-wise motion vectors. 
The motion condition then inputs to a motion modulation ($\mathcal{M}^2$) layer in which it is separately mapped to a pair of transformation parameters $\beta$ and $\gamma$. Two groups of $1\times1$ convolutional layers are independently adopted for generating each of the parameters $\beta$ and $\gamma$. The low-level RGB features from the appearance network are adjusted by $\beta$ and $\gamma$. The motion modulation layer is capable to be added to any bottom layer of the appearance network, including conv1, conv2, conv3 and conv4. All of them share the motion condition layer. The whole network is end-to-end trainable.

\begin{figure}[t!]
\centering
\includegraphics[scale=0.52]{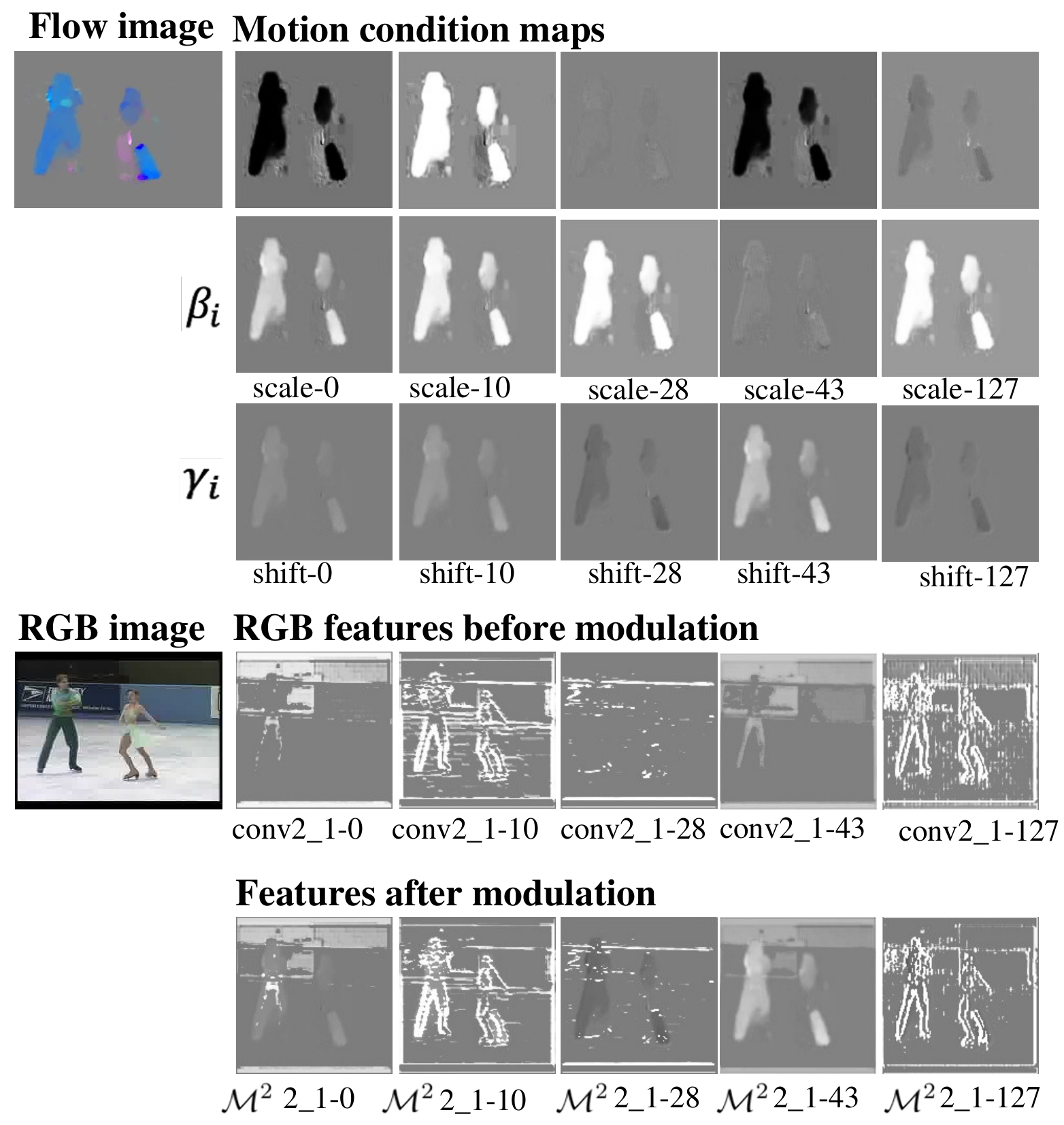}
\caption{\textbf{Feature maps.} Visualization of the motion condition maps, scale maps, shift maps, RGB features without modulation and features with modulation. The modulated features focus more on moving actors.}
\label{Fig:maps}
\vspace{-10pt}
\end{figure}

{\bf{Feature visualization.}}  In order to intuitively understand the method, we show the generated feature maps from the appearance network before and after modulation by motion condition in Figure ~\ref{Fig:maps}. We randomly select some feature maps from the motion condition layer in the first row. The features are low-level and sparse, which are taken as prior conditions. From the second row to the last row, we show the corresponding scale ($\beta$) and shift ($\gamma$) maps generated from conditions, RGB features without modulation and features modulated by $\beta$ and $\gamma$. It is interesting to see the difference between the features without and with modulation in Figure~\ref{Fig:maps}. For example the modified features of the actor areas in feature maps 0 and 43, after modulation, especially for the female ice skater, which is blended into the background on the regular RGB stream. On the 28-th feature map, a feature response is even hard to see on both actors before modulation. Feature maps 10 and 127 show the change in $x$-direction features and $y$-direction features. The flow condition pushes the model to focus on moving actors.


{\bf{Training loss.}} In order to demonstrate the generalization and flexibility of the proposed method, we embed the motion condition layer and the motion modulation layer in a single-frame appearance stream and a multi-frame appearance stream. The basic loss function is derived from the one for object detection~\cite{liu2016ssd,ren2015faster}. Defining ${x^p}_{ij}=\{1,0\}$ as an indicator for matching the $i$-th default box to the $j$-th ground truth box of action category $p$. The overall loss function contains the localization ($loc$) loss and the confidence ($conf$) loss:
\begin{equation}
\mathcal{L}(x,c,l,g) = \frac{1}{N}(\mathcal{L}_{loc}(x,l,g)+\mathcal{L}_{conf}(x,c))
\end{equation}
with $N$ representing the number of matched default boxes. $c$ represents multiple classes confidences. $l$ and $g$ are the predicted box and the ground truth box.

The confidence loss applies the softmax loss as below:
\begin{equation}
\begin{aligned}
\mathcal{L}_{conf}(x,c) &= -\sum_{i\in{Pos}}^{N}x_{ij}^{p}log(\hat{c}_{i}^{p})-\sum_{i\in{Neg}}log(\hat{c}_{i}^{0}) \\
\hat{c}_{i}^{p} &=\frac{exp(c_{i}^{p})}{\sum_{p}exp(c_{i}^{p})}
\end{aligned}
\end{equation}
The localization loss applies a smooth L1 loss~\cite{girshick2015fast} between the predicted box and the ground truth box. The network regresses to offsets for the center ($cx,cy$) of the default box($d$) and for its width ($w$) and height ($h$).
\begin{equation}
\begin{aligned}
\mathcal{L}_{loc}(x,l,g) &=\sum_{i\in{Pos}}^{N}\sum_{m\in\{cx,cy,w,h\}}x_{ij}^{k}smooth_{\mathrm{L}1}(l_{i}^{m}-\hat{g}_{j}^{m}) \\
\hat{g}_{j}^{cx} &= (g_{j}^{cx}-d_{i}^{cx})/d_{i}^{w} \qquad
\hat{g}_{j}^{cy} = (g_{j}^{cy}-d_{i}^{cy})/d_{i}^{h} \\
\hat{g}_{j}^{w} &= log (\frac{{g}_{j}^{w}}{d_{i}^{w}}) \qquad \hat{g}_{j}^{h} = log (\frac{{g}_{j}^{h}}{d_{i}^{h}})
\end{aligned}
\end{equation}
For the multi-frame appearance stream, we follow Kalogeiton \etal~\cite{kalogeiton2017action} to train the network.

{\bf{Two-in-one two-stream.}} Our method emphasizes to utilize RGB and optical flow information in one stream. Furthermore, it is possible to follow the standard practice of two-stream action detection. We train a two-in-one detector conditioned on flow images, and a separate flow detector which only takes as input the flow images. 
For a single-frame two-in-one two-stream, we use average fusion method to merge the results from each stream following~\cite{singh2017online}. And for multi-frame two-stream, the late fusion ~\cite{feichtenhofer2016convolutional} is a better choice ~\cite{kalogeiton2017action}.

{\bf{Linking.}} Once the frame-level detections or tubelet detections are achieved, we link them to build action tubes. We adopt the linking method described in~\cite{singh2017online} for frame-level detections and the method in~\cite{kalogeiton2017action} for tubelet detections.

Code is available at \href{https://github.com/jiaozizhao/Two-in-One-ActionDetection}{https://github.com/jiaozizhao/Two-in-One-ActionDetection}.

\section{Experiments}
\subsection{Datasets, Metrics \& Implementation}
{\bf{Datasets.}} We perform experiments on three action detection datasets.
{\it{UCF101-24}}~\cite{soomro2012ucf101} is a subset of {\it{UCF101}}. It contains 24 sport classes in 3207 untrimmed videos. Each video contains a single action category. Multiple action instances with the same class, but different spatial and temporal boundaries may occur. We use the revised annotations for {\it{UCF101-24}} from~\cite{singh2017online}. 
{\it{UCF-Sports}}~\cite{rodriguez2008action} contains 10 sport classes in 150 trimmed videos. We follow ~\cite{lan2011discriminative} to divide the training and test splits.
{\it{J-HMDB}}~\cite{jhuang2013towards} contains 21 action categories in 928 trimmed videos. We report the average results on three splits.

{\bf{Metrics.}} Following ~\cite{soomro2016predicting,weinzaepfel2015learning,saha2016deep}, we utilize video mean Average Precision ($mAP$) to evaluate action detection accuracy.
We calculate an average of per-frame Intersection-over-Union (IoU) across time between tubes.
A detection is correct if it's IoU with the ground truth tube is greater than a threshold and its action label is correctly assigned. We compute the average precision for each class and report the average over all classes.

{\bf{Implementation.}}
We adopt a real-time single shot multibox detector (SSD) network~\cite{liu2016ssd} as the backbone. We insert the developed motion layers into two state-of-the-art appearance SSD networks, one based on single frame~\cite{singh2017online} and the other based on multiple frames~\cite{kalogeiton2017action}. We use VGG-16 pre-trained weights on ImageNet as model initialization. The input size is 300x300 for both of them. We follow~\cite{kalogeiton2017action} to use 6 continuous frames as input to the multi-frame SSD. The initial learning rate is set to 0.001 for the single-frame network and 0.0001 for the multi-frame network on all the three datasets and changed by applying step decay strategy. We trained a flow-stream, an RGB-stream and our two-in-one stream for 13.2, 13.2 and 15.5 hours, respectively.

Alternatively, we considered to use appearance information to modulate flow stream. However, it does not work well. It appears difficult to modulate features from flow images which are sparse, using RGB images which are more dense.


\begin{table*}[t]
	\centering
\scalebox{0.9}{
		\begin{tabular}{cccc|ccc}
		\toprule
        \multicolumn{4}{c|}{\textbf{Action Detection}} &  \multicolumn{3}{c}{\textbf{Action Classification}}\\
         \cmidrule(lr){1-4} \cmidrule(lr){5-7}
		Method & {\textit{mAP}} & \multicolumn{2}{c|}{\textit{Efficiency}} & {\textit{Top1 Accuracy}} & \multicolumn{2}{c}{\textit{Efficiency}}\\
         \cmidrule(lr){2-2} \cmidrule(lr){3-4} \cmidrule(lr){5-5} \cmidrule(lr){6-7}
			& \% & sec/frame & \# param. (M) & \% & sec/frame & \# param. (M) \\

			\midrule
          flow-stream  & 11.60 & 0.04 & 26.82 & 81.65 & 1.10 & 58.35  \\
          RGB-stream  & 18.49 & 0.04 & 26.82 & 84.99 & 1.10 & 58.35  \\
           two-stream  & 19.79 & 0.09 & 53.64 & 91.14 & 2.10 & 116.70 \\
           two-in-one stream  & 20.15 & 0.04 & 26.93 & 86.94 & 1.15 & 58.48  \\
           two-in-one two stream  & \bf{22.02} & 0.09 & 53.75 & \bf{92.00} & 2.13 & 116.83  \\
	\bottomrule
		\end{tabular}
		}
	\caption{\textbf{Two-in-one \textit{vs.} baselines} for action detection on {\it{UCF101-24}} and action classification on {\it{UCF101}}. Two-in-one with motion modulation works well for both action detection and action classification.}
	\label{tab: Tab1}
		\vspace{-10pt}
\end{table*}

\subsection{Ablation Study}
All the ablation studies are performed on {\it{UCF101-24}}. We only report $mAP$ at the most challenging high $IoU$ thresholds 0.5:0.95 (with step 0.05). Initially, in order to maintain the spatial pixel-wise motion vectors, we apply 1x1 convolution kernels to all layers in the motion condition layer and the motion modulation layer. We use layer parameter stride to control the size of $\beta$ and $\gamma$. Then the motion modulation layer is applied to conv1 of SSD. Flow images are generated using the method in~\cite{brox2004high}, which we refer to as BroxFlow.

{\bf{Two-in-one \textit{vs.} baselines.}} We compare the two-in-one stream to its corresponding RGB-stream, flow-stream and two-stream in Table~\ref{tab: Tab1}. Runtime and \# param. are also reported for comparing the efficiency. Our single two-in-one stream exceeds a single RGB-stream by $1.5\%$. Notably, two-in-one even outperforms the corresponding two-stream with only half the computation cost and \# param..

We also consider action classification, on {\textit{UCF101}}. We follow~\cite{wang2015towards}, with ResNet152 as backbone. The {\textit{Top 1}} accuracy and efficiency shown in Table~\ref{tab: Tab1} illustrate our strategy also works for action classification and generalizes beyond SSD with VGG16. For training, our two-in-one stream converges at the 100-th epoch, but the RGB- and flow-stream converge at 200-th and 300-th epoch, respectively. Our motion modulation strategy works better for the detection task, which needs localization representations that are
translation-variant, compared to the classification task which favors translation invariance.

\begin{figure}[t]
\centering
\includegraphics[scale=0.11]{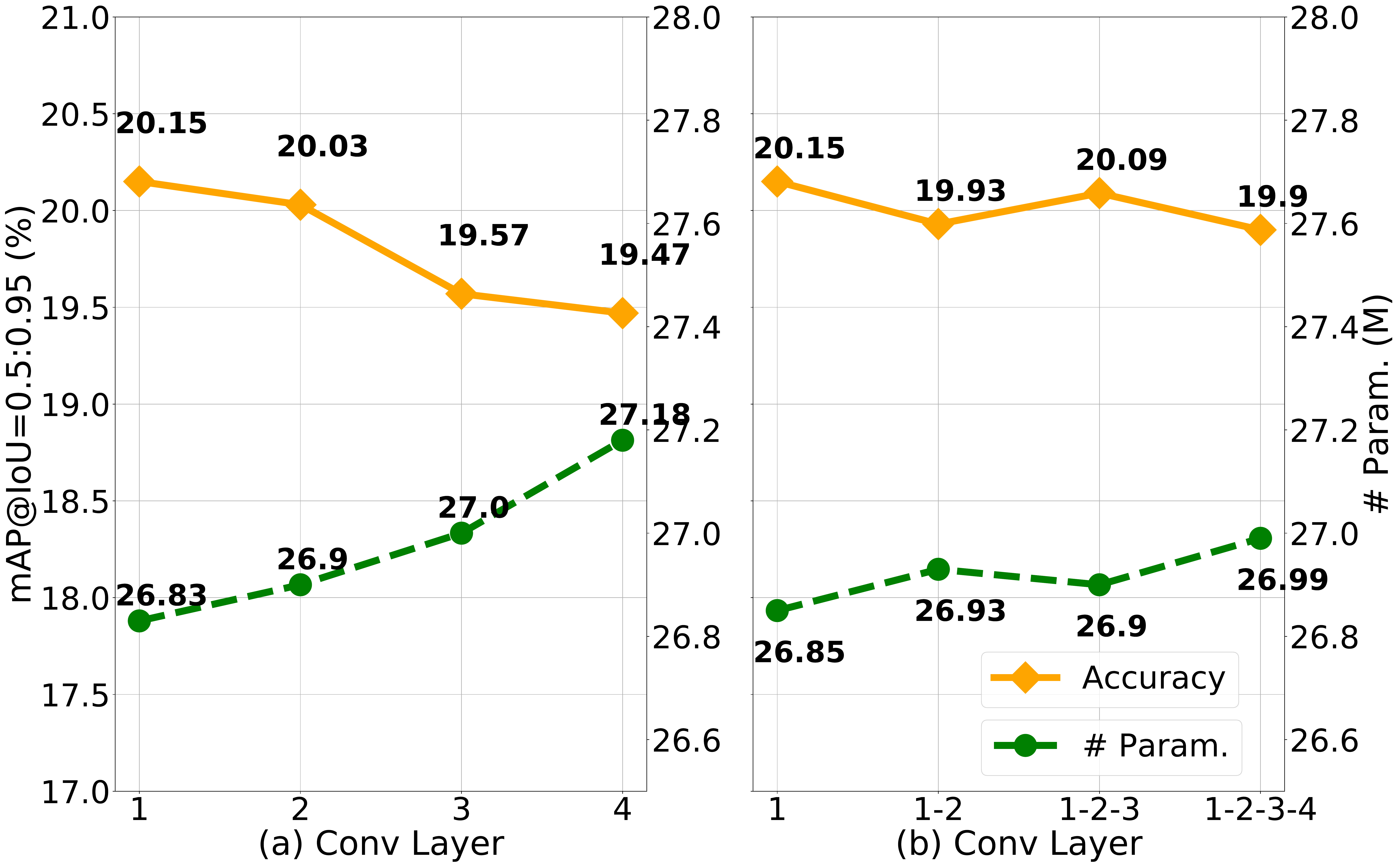}
\caption{\textbf{Where to add the modulation layer?} Accuracy on {\it{UCF101-24}} and \# param. with varying: (a) single modulated layer, and (b) multiple modulated layers. A single modulation layer at conv1 gets the best result. }
\label{Fig:c1}
\end{figure}
\begin{figure}[t]
\centering
\includegraphics[scale=0.4]{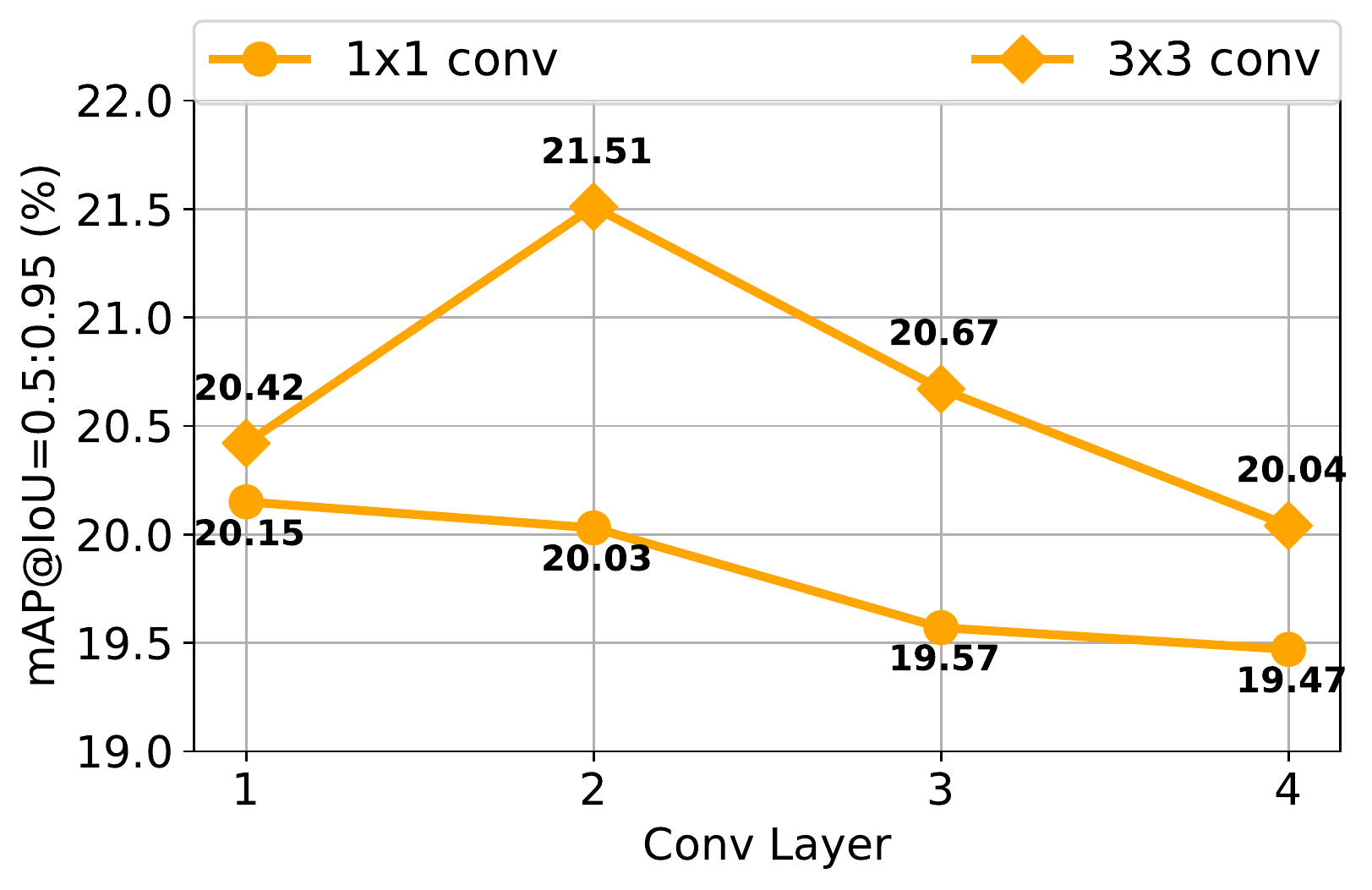}
\caption{\textbf{How to design the condition layer?} Comparing accuracy on {\it{UCF101-24}} when applying 1x1 conv or 3x3 conv to the last layer of the motion condition layer. The 3x3 conv performs better.}
\label{Fig:c2}
\vspace{-10pt}
\end{figure}

{\bf{Where to add the modulation layer?}} The motion condition layer is leveraged to generate low-level motion features as flow images are more sparse. We add the motion modulation layer to the bottom convolutional layers with low-level RGB features. We conduct two experiments on which layer to add the modulation. We compare the accuracy and \# param. after applying a modulation layer to conv1, conv2, conv3 and conv4 in Figure~\ref{Fig:c1} (a). Accuracy decreases and \# param. increases slightly for deeper layers. 
Next we add the modulation layers to multiple convolutional layers simultaneously in Figure~\ref{Fig:c1} (b). Applying multiple modulation layers does not change the results much. Thus, we prefer to use a single modulation layer. Note that accuracy drops for deeper layers as we use 1x1 convolution kernels to process flow images, leading to smaller receptive field for deeper layers.
%

{\bf{How to design the condition layer?}} To further improve the method, we consider whether the 1x1 convolution kernel for the motion condition layer is the best choice. Besides keeping the spatial pixel-wise motion, it may need to consider some context of motion to better fit the RGB features. We adopt the 3x3 convolution kernels to the last layer of the condition network. Figure~\ref{Fig:c2} demonstrates that considering motion context boosts the accuracy for all layers. As a bigger receptive field is used, the conv2 model achieves the best results, about $1.5\%$ improvement compared to 1x1 convolution kernels. The run time hardly increases for deeper layers, and is still 0.04 sec per frame. The \# param. are 26.85, 26.92, 27.01 and 27.19 M respectively of conv1, conv2, conv3 and conv4. Considering the trade-off between the results and parameters, we believe conv2 provides the best accuracy/efficiency trade-off.


\begin{figure}[t]
\centering
\includegraphics[scale=0.3]{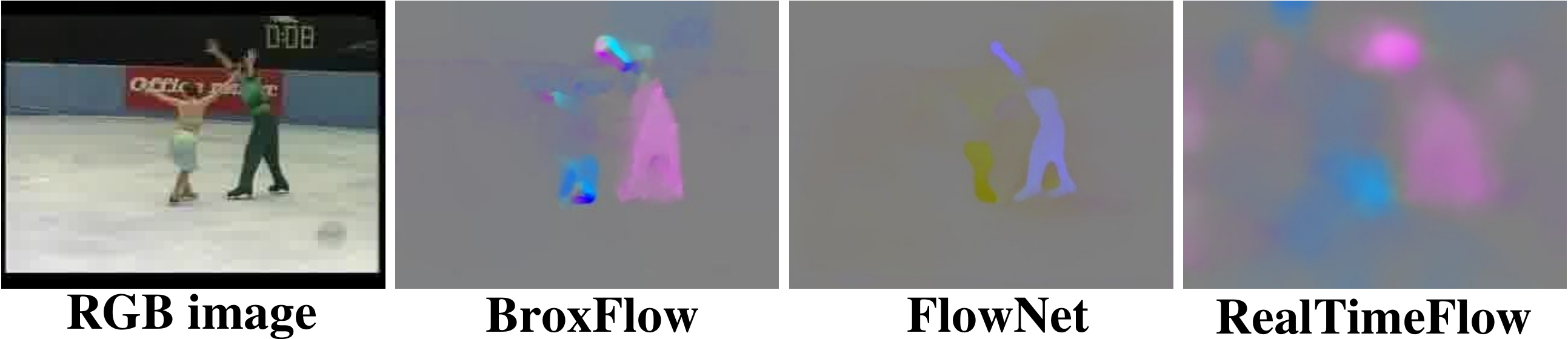}
\caption{\textbf{What flow?} Examples of flow images generated by different flow methods.}
\label{Fig:f1}

\end{figure}

\begin{table}[t!]
	\renewcommand\arraystretch{1.1}
	\centering
	\resizebox{\columnwidth}{!}{%
		\begin{tabular}{lrrr}
		\toprule
		& {\textbf{BroxFlow}} & {\textbf{FlowNet}} &{\textbf{RealTimeFlow}} \\
			\midrule
           flow-stream  & 11.60  & 7.13 & 3.58\\
           RGB-stream   & 18.49 & 18.49  & 18.49 \\
           two-stream   & 19.79 & 19.75  & 18.53\\
           two-in-one stream  & {\bf{21.51}}& 19.97 & 19.16\\
		\bottomrule
		\end{tabular}
		}
	\caption{\textbf{What flow?} No matter what flow images are applied on {\it{UCF101-24}}, our two-in-one stream outperforms the corresponding flow-, RGB- and two-stream. We obtain the best result with BroxFlow.}
	\label{tab:Tab3}
\end{table}


\begin{figure}[t!]
\centering
\includegraphics[scale=0.2]{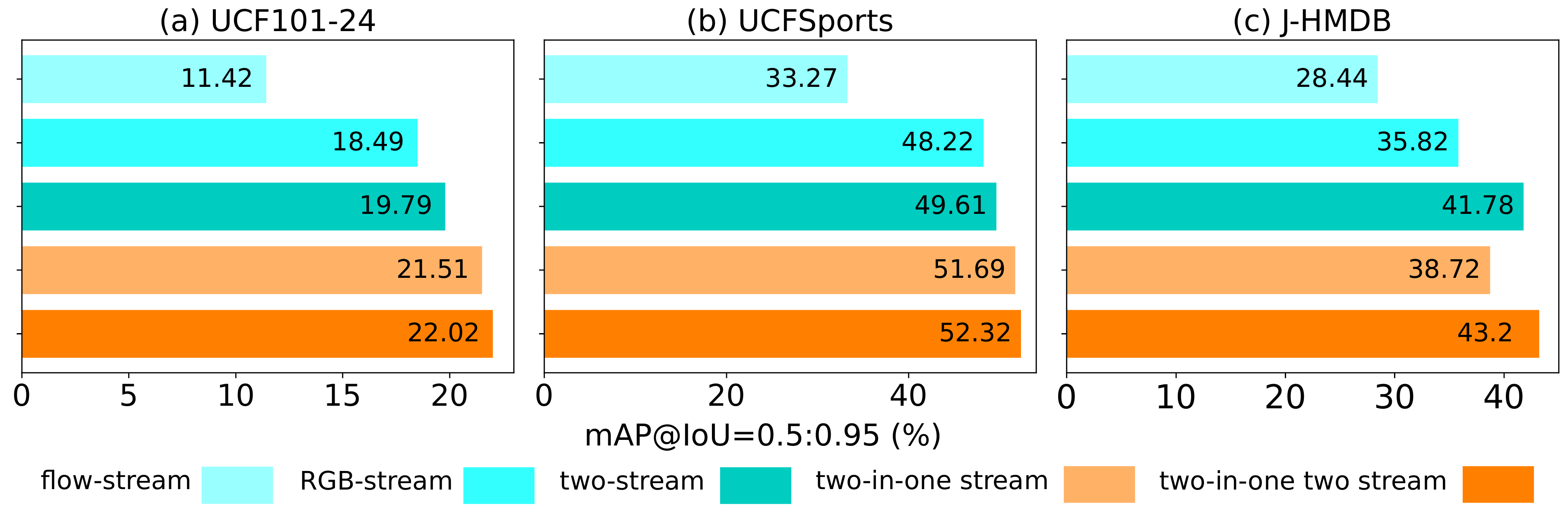}
\caption{\textbf{Generalization ability.} Accuracy comparison on: (a) {\it{UCF101-24}}, (b) {\it{UCFSports}}, (c) {\it{J-HMDB}}, with different methods. Two-in-one stream even outperforms two-stream on {\it{UCF101-24}} and {\it{UCFSports}}. Two-in-one stream fused with a flow-stream obtains the best accuracy on all three datasets.}
\label{Fig:bar}
\vspace{-10pt}
\end{figure}


{\bf{What flow?}} As we leverage flow information as prior conditions, we wonder how the model is influenced by flow images. Here we adopt flow images generated by three different methods (seen in Figure~\ref{Fig:f1}) and evaluate how our strategies work. We use BroxFlow~\cite{brox2004high} (accurate flow method), Flownet~\cite{dosovitskiy2015flownet} (deep network method) and a real-time but less accurate optical flow method~\cite{kroeger2016fast} (RealTimeFlow). From Table~\ref{tab:Tab3}, it is concluded that no matter which kind of flow images are applied, our two-in-one stream outperforms RGB-streams and corresponding two-streams. We also note that the more accurate the flow images, the more improvement the two-in-one stream obtains. Even when using the somewhat noisy RealTimeFlow images, the two-in-one stream still improves the RGB-stream. However, a two-stream based on RealTimeFlow obtains almost the same accuracy as the RGB-stream, which illustrates that two-stream depends on the the quality of flow images. Our two-in-one stream is more robust to the quality of flow images. Moreover, we report the flow computation in seconds/frame for the three kinds of flow methods: BroxFlow (0.098), FlowNet (0.183) and RealTimeFlow (0.014). RealTimeFlow only needs 0.014 seconds to generate one flow image, at the expense of a slightly lower {\textit{mAP}}.

{\bf{Generalization ability.}} To stress the generalization ability of our proposal, we compare the results on three different datasets. Following the conclusions of our ablation so far, we use the BroxFlow image for generating condition and apply a 3x3 kernel to the last layer of the motion condition layer. The motion modulation layer is only leveraged for the conv2 layer of the appearance stream.
%
%
We report results in Figure~\ref{Fig:bar}.

Obviously, the proposed two-in-one stream performs better than other one-stream networks. It is noteworthy that our two-in-one stream even outperforms traditional two-stream networks on {\it{UCF101-24}} and {\it{UCFSport}} by 2\% with only half the parameters of a two-stream network. On {\it{J-HMDB}}, two-in-one is 3\% higher than RGB-stream but 3\% lower than two-stream. We look into {\it{J-HMDB}} and find that most videos in the dataset have neighbouring repeated frames. For fair comparison, we just download the BroxFlow images used in ~\cite{singh2017online, kalogeiton2017action}. However, the provided BroxFlow image between the two repeated RGB frames is not 0, as it should be, but similar to the last flow frame.
%
%
The issue affects our two-in-one stream due to the fact that we need correct flow image as the condition of the corresponding RGB frame. We expect that two-in-one will present better results on {\it{J-HMDB}} after correcting the flow images. As expected, adding a separate flow-stream to our two-in-one stream gives the best accuracy on all datasets.


\begin{figure}[t!]
\centering
\includegraphics[scale=0.55]{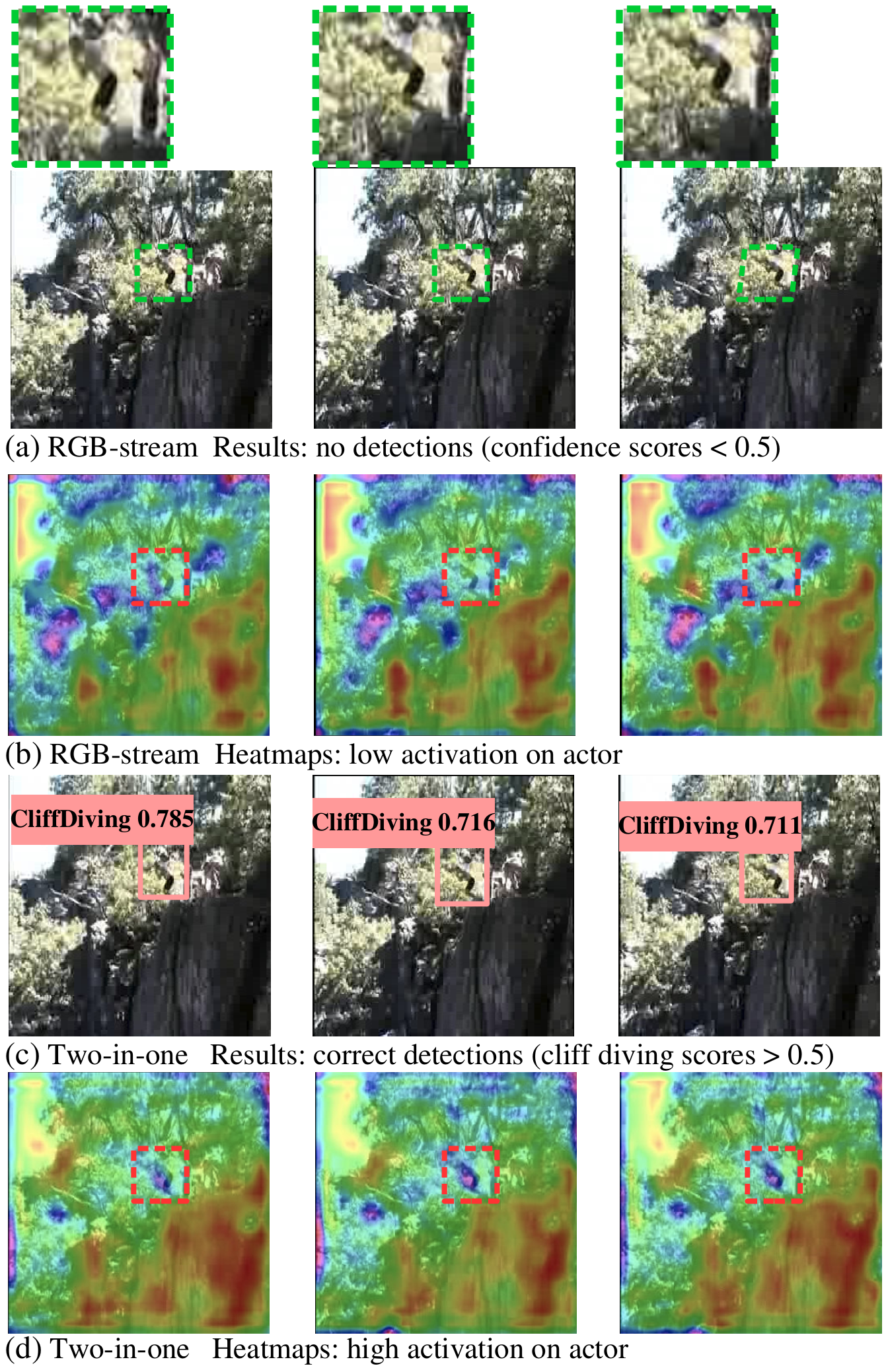}
\caption{\textbf{Visualization of detection and heatmaps} on conv4 layers from RGB-stream network in (a) (b) and two-in-one stream network in (c) (d). We add the green dashed boxes to indicate the action. The two-in-one stream has higher activation on actions, resulting in correct detection.}
\label{Fig:heatmap}
\vspace{-10pt}
\end{figure}

\subsection{Qualitative Analysis}
The motion condition layer and the motion modulation layer are beneficial to generate better video representations for spatio-temporal action detection. But how do the layers make a difference to the appearance network? To understand this behavior, we visualize in Figure~\ref{Fig:heatmap}, the detection results of an RGB-stream network and a two-in-one network. Also, we visualize the gradient-weighted class activation heatmaps~\cite{zhou2016learning} for better understanding how the motion conditions influence the behavior of the appearance network. We choose a challenging case of cliff diving here. The image resolution is low and the actor is quite tiny. The cluttered background obviously increases the difficulty to detect actions. We manually overlay green dashed boxes to indicate the locations of the actor and zoom in to highlight where the action is happening. The second row shows that the RGB-stream fails to detect any actions. From the corresponding heatmaps, it is apparent that the appearance network pays more attention to the background than to the actions. There are only weak responses on the action positions. We manually overlay red dashed boxes to highlight the position of the actor on the heatmaps. From the heatmaps for the two-in-one network in the last row, we clearly see it is capable to balance the activation on actions and background. The responses on action positions are strengthened. As expected, the two-in-one stream performs better than the RGB-stream. It outputs correct detections for cliff diving on all the frames (forth row ).


\begin{table*}[t!]
	\renewcommand\arraystretch{1}
	\centering
	\scalebox{0.9}{
		\begin{tabular}{llccccccccc}
		\toprule
	  & \multicolumn{4}{c}{\textit{UCF101-24}} & \multicolumn{3}{c}{\textit{UCFSports}} &\multicolumn{3}{c}{\textit{J-HMDB}}   \\
		\cmidrule(lr){2-5} \cmidrule(lr){6-8} \cmidrule(lr){9-11}
		& {{0.20}}& {{0.50}}& {{0.75}}& {{0.50:0.95}}  & {{0.50}}& {{0.75}}& {{0.50:0.95}} & {{0.50}}& {{0.75}}& {{0.50:0.95}} \\

		\midrule
\rowcolor{Gray}
        \textbf{Single-frame} & & & & & & & & & &\\
           Peng \& Schmid~\cite{peng2016multi}  & 71.80 & 35.90 & 1.60 & 8.80 & \bf{94.80} & 47.30 & 51.00 & 70.60 & \it{48.20} & \it{42.20}  \\
           Saha \etal~\cite{saha2016deep} & 66.70 & 35.90 & 7.90 & 14.40 &  --&  -- &  -- & \it{71.50} & 43.30 & 40.00  \\
           Behl \etal~\cite{behl2017incremental}   & 71.53 & 40.07 & 13.91 & 17.90 & -- & --& -- & --& -- & --  \\
           Singh \etal~\cite{singh2017online}  & 73.50 & 46.30 & 15.00 & 20.40 & --& --& --& \bf{72.00} & 44.50 & 41.60  \\
           \it{This paper: Two-in-one}  & \it{75.13} & \it{47.47} & \it{17.21} & \it{21.51}  & 87.46 & \it{57.81} & \it{51.69} & 60.99 & 47.23 & 38.72  \\
           \it{This paper: Two-in-one two stream}  & \bf{77.49} & \bf{49.54} & \bf{17.62} & \bf{22.02}  & \it{87.81} & \bf{62.67}& \bf{52.32} & 70.00 & \bf{52.00} & \bf{43.20}  \\
\midrule
\rowcolor{Gray}
\textbf{Multi-frame} & & & & & & & & & &\\
			Saha \etal~\cite{saha2017amtnet} & 63.06 & 33.06 & 0.52 & 10.72 & --  & -- & -- & 57.31 & --& --  \\
			Kalogeiton \etal~\cite{kalogeiton2017action} & 76.50 & 49.20 & 19.70 & 23.40 & 92.70 & 78.40 & 58.80 & \it{73.70} & \it{52.10} & \it{44.80}\\
			Singh \etal~\cite{singh2018tramnet}  & \bf{79.00} & \bf{50.90} & 20.10 & \it{23.90} & --  & --& --  & --& --  & --   \\
			\it{This paper: Two-in-one} & 75.48 & 48.31 & \it{22.12} & \it{23.90} & \it{92.74} & \it{83.64} & \it{59.60} & 57.96 & 42.78 &  34.56   \\
			 \it{This paper: Two-in-one two stream}  & \it{78.48} & \it{50.30} & \bf{22.18} & \bf{24.47}  & \bf{96.52} & \bf{90.41} & \bf{63.59} & \bf{74.74} & \bf{53.28} & \bf{45.01} \\
			\bottomrule
		\end{tabular}
		}
	\caption{\textbf{Accuracy comparison to the state-of-the-art.} Bold means top accuracy and italic means second top accuracy. For the high overlap setting of $mAP@IoU$=0.5:0.95, our two-in-one stream works well in both a single-frame and multiple-frame network for all three datasets. When we add an additional flow-stream to obtain a two-in-one two stream we further improve accuracy.}
	\label{tab: Tab4}
	\vspace{-10pt}
\end{table*}

\begin{figure}[t!]
\centering
\includegraphics[scale=0.26]{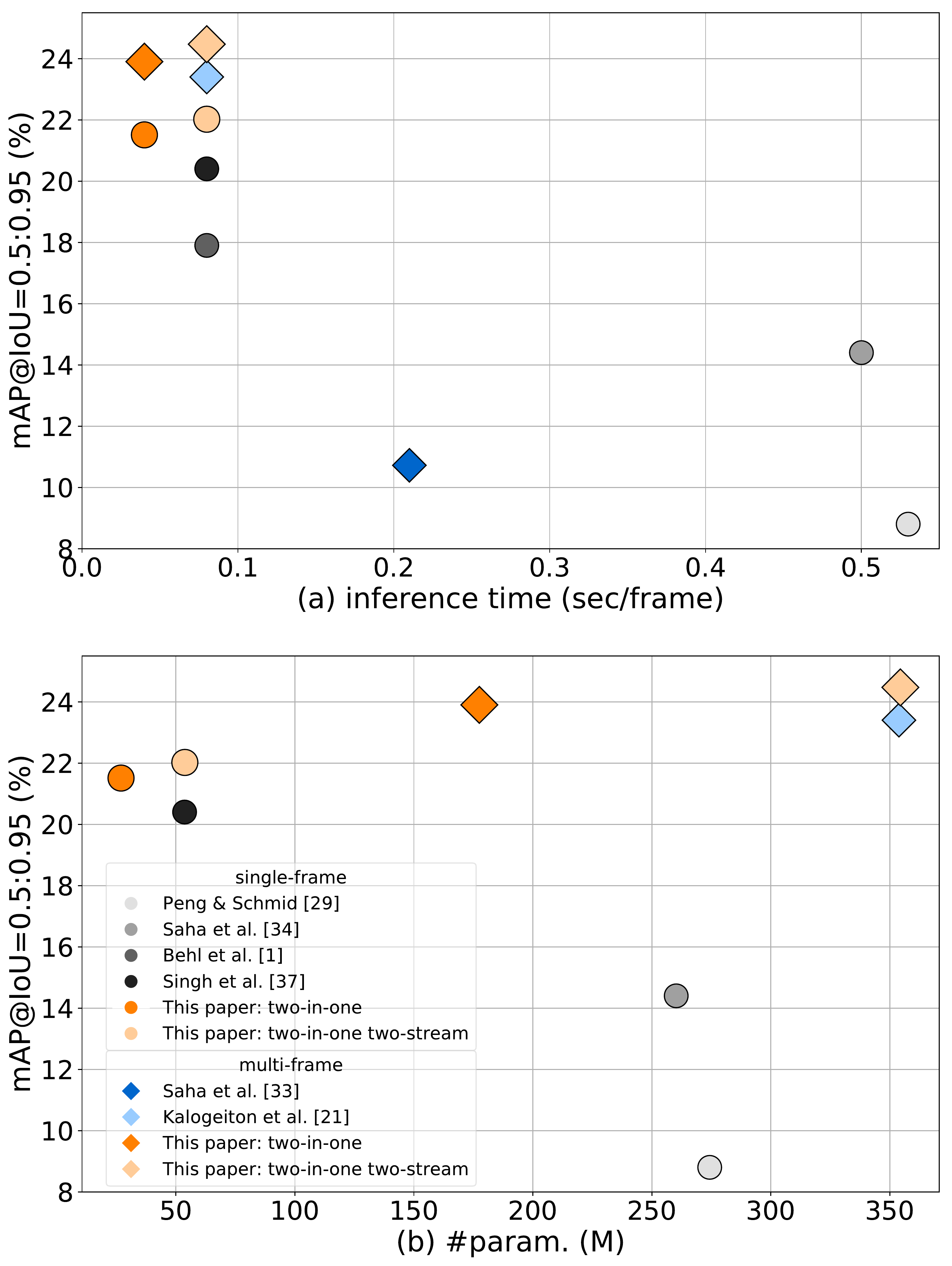}
\caption{\textbf{Efficiency comparison to the state-of-the-art.} Accuracy \textit{vs.} (a) inference time (second per frame) and (b) \# param. (M) on {\it{UCF101-24}}. Our two-in-one stream best balances accuracy and efficiency.}
\label{Fig:PvsT}
\vspace{-10pt}
\end{figure}

\subsection{Comparison to the State-of-the-art}

{\bf{Accuracy.}}
For fair comparisons, we just use the original images as in all the state-of-the-arts, without camera motion removal. We compare the $mAP$ at variable $IoU$ thresholds in Table~\ref{tab: Tab4}. Considering the most challenging high $IoU$ thresholds 0.5:0.95, we observe that for the single-frame setting, our two-in-one stream achieves even better results than existing two-stream methods on {\it{UCF101-24}} and {\it{UCFSports}}. For instance, two-in-one stream outperforms Singh \etal~\cite{singh2017online} with the same SSD detector by more than $1\%$ and Peng and Schmid~\cite{peng2016multi} with a Faster-RCNN detector by an absolute $12\%$ on {\it{UCF101-24}}. As analyzed previously, two-in-one stream performs modest on {\it{J-HMDB}} because of the data issue of the provided BroxFlow images. When we combine two-in-one into a regular two-stream network by fusing with a flow-stream, it produces good results on all three datasets. Compared to two-in-one stream, it gets about $5\%$ improvement on {\it{J-HMDB}}. Moreover, when feeding our two-in-one network variants with multiple frames, as suggested by Kalogeiton \etal~\cite{kalogeiton2017action}, our two-in-one stream outperforms the two-stream ~\cite{kalogeiton2017action} a little  on {\it{UCF101-24}} and {\it{UCFSports}} with only half computation and the number of parameters. Our two-in-one stream fused with a flow stream further boosts the results, outperforming the very recent work of Singh \etal~\cite{singh2018tramnet}.

{\bf{Efficiency.}} Besides good detection accuracy, our method has the advantage of a reduced inference time and less \# param.. Here we compare our methods from the efficiency aspect to the state-of-the-art on {\it{UCF101-24}}. We test our models on one NVIDIA GTX 1080 GPU. The trade-off between accuracy and inference time, as well as parameters are visualized in Figure~\ref{Fig:PvsT}. Among the single-frame methods, our two-in-one stream has the fastest run time with 0.04s per frame, two times faster than~\cite{behl2017incremental} and~\cite{singh2017online} and much faster than~\cite{saha2016deep} and~\cite{peng2016multi} (about 0.5s per frame). Moreover, the \# param. of our two-in-one stream is smallest, about 26.93 M. While our two-in-one accuracy is even better than the two-stream methods by~\cite{behl2017incremental,saha2016deep,peng2016multi,singh2017online}. Combining our two-in-one stream with a standard flow-stream gains an accuracy improvement at the expense of more computation and parameters. Our two-in-one alternative even outperforms~\cite{kalogeiton2017action} a little in accuracy with only half the parameters. The two-in-one two stream further improves the result with almost similar inference time, but slightly more parameters.
We conclude that two-in-one stream networks provide a good accuracy/efficiency trade-off.

\section{Conclusion}
We propose an effective and efficient two-in-one stream network for spatio-temporal action detection. It takes flow images as prior motion condition when training an RGB-stream network. The network's motion condition layer and motion modulation layer address two issues in action detection: frame-level RGB images lack motion information and (static) background-context may dominant the learned representation. Our two-in-one stream achieves state-of-the-art accuracy at high $IoU$ thresholds, using only half of the parameters and computation of two-stream alternatives. Besides motion, we believe that other information such as depth-maps or infrared images may help locate the actors, and can be exploited as additional prior conditions for training two-in-one streams.

{
	\small
	\textbf{Acknowledgments}
	Supported by the Intelligence Advanced Research Projects Activity (IARPA) via Department of Interior/Interior Business Center (DOI/IBC) contract number D17PC00343. The U.S. Government is authorized to reproduce and distribute reprints for Governmental purposes notwithstanding any copyright annotation thereon. Disclaimer: The views and conclusions contained herein are those of the authors and should not be interpreted as necessarily representing endorsements, either expressed or implied, of IARPA, DOI/IBC, or the U.S. Government.
}

{\small
\bibliographystyle{ieee_fullname}
\bibliography{egpaper_final}
}

\end{document}